%% file: main.tex
\documentclass[10pt,twocolumn,letterpaper]{article}

\usepackage[pagenumbers]{cvpr} 

\input{preamble}
\definecolor{cvprblue}{rgb}{0.21,0.49,0.74}
\usepackage{placeins}

\usepackage[pagebackref,breaklinks,colorlinks,allcolors=cvprblue]{hyperref}
\usepackage{multirow}
\usepackage[table]{xcolor} 
\usepackage{hyperref}  

\usepackage{cuted} 
\usepackage{caption}
\definecolor{darkgreen}{RGB}{0,100,0} 
\usepackage{float}  
\usepackage{xcolor}   
\usepackage{pifont}   

\newcommand{\cmark}{\textcolor{darkgreen}{\ding{51}}} 
\newcommand{\xmark}{\textcolor{red}{\ding{55}}}   

\def\paperID{4032} 
\def\confName{CVPR}
\def\confYear{2026}

\title{Unifying Scientific Communication: Fine-Grained Correspondence Across Scientific Media}

\author{
Megha Mariam K.M\\
IIIT Hyderabad\\
{\tt\small megha.km@research.iiit.ac.in}
\and
Vineeth N. Balasubramanian\\
Microsoft Research India \& IIT Hyderabad\\
{\tt\small vineeth.nb@microsoft.com}
\and
\and
C.V. Jawahar\\
IIIT Hyderabad\\
{\tt\small jawahar@iiit.ac.in}
}

\begin{document}
\maketitle

\begin{strip}
    \vspace{-1.5cm}
  \centering
  \includegraphics[width=\textwidth]{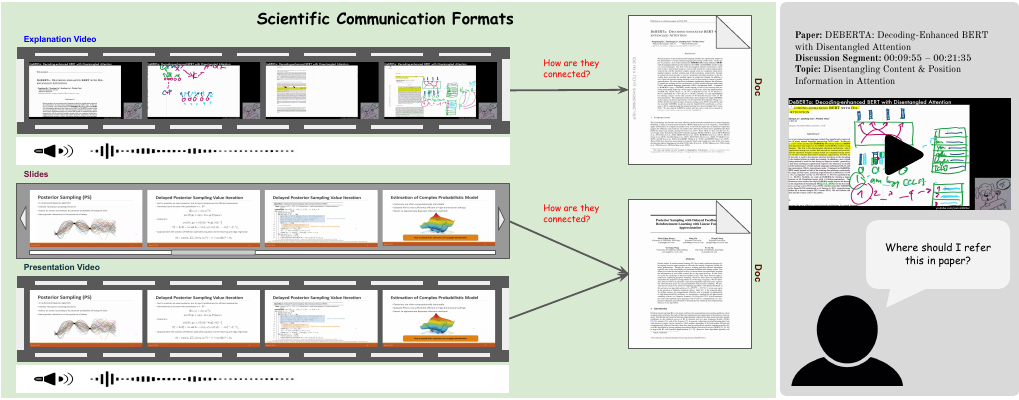}
  \captionof{figure}{\textbf{Scientific communication formats and their interconnections:} 
The figure illustrates how research knowledge is represented and shared across multiple formats, including research papers (Docs), presentation slides, conference videos, and explanation videos. 
These formats are not isolated; rather, strong semantic and structural connections exist between them. 
Slides often summarize and visualize key insights from papers, presentation videos provide verbal and contextual elaboration, and explanation videos further distill the content for broader understanding. 
Together, they form a coherent, interconnected network of scientific communication that captures complementary aspects of the same underlying research.}
  \label{fig:intro}
\end{strip}

\input{sec/0_abstract}    

\input{sec/1_intro}

\input{sec/2_relWork}

\begin{figure*}[t]
  \centering
   \includegraphics[width=\linewidth]{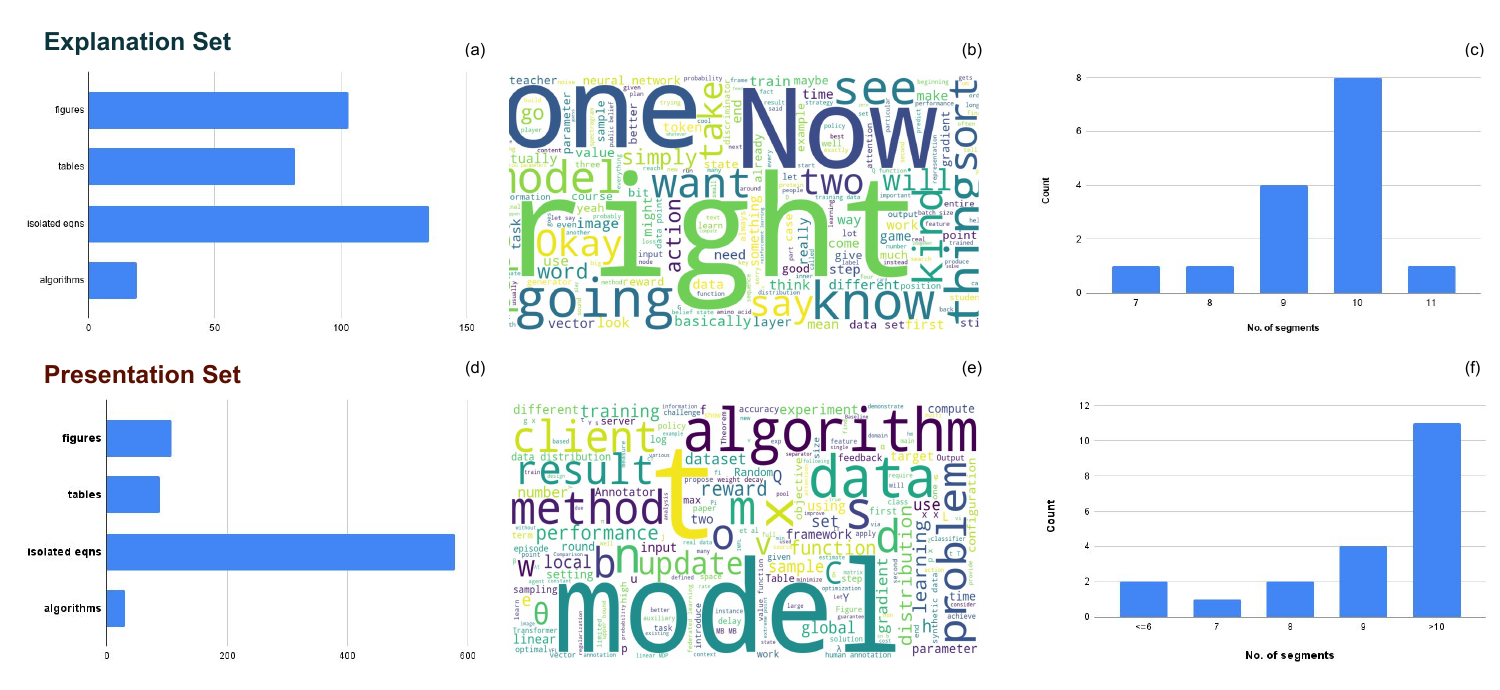}

   \caption{\textbf{Statistics of the Explanation and Presentation sets:} (a) and (d) show the distribution of algorithms, equations, tables, and figures in the papers; (b) presents the word cloud generated from ASR transcripts; (e) presents the word cloud generated from slide text and ASR transcripts; (c) and (f) depict the number of videos per segment category.}
   \label{fig:stats}
   \vspace{-0.4cm}
\end{figure*}
\input{sec/3_Dataset}
\input{sec/4_probdefn}

\input{sec/5_baselines}

\input{sec/7_res}

\input{sec/8_conclusion}

\section*{Acknowledgments}
This work is supported by the MeitY, Government of India, through the NLTM Bhashini project (\url{https://bhashini.gov.in}). We sincerely thank the anonymous reviewers for their valuable feedback, which helped improve the quality of this paper.
 
{
    \small
    \bibliographystyle{ieeenat_fullname}
    \bibliography{main}
}


\input{suppl}
\end{document}

%% file: sec/0_abstract.tex
\begin{abstract}
The communication of scientific knowledge has become increasingly multimodal, spanning text, visuals, and speech through materials such as research papers, slides, and recorded presentations. These different representations collectively convey a study’s reasoning, results, and insights, offering complementary perspectives that enrich understanding. However, despite their shared purpose, such materials are rarely connected in a structured way. The absence of explicit links across formats makes it difficult to trace how concepts, visuals, and explanations correspond, limiting unified exploration and analysis of research content.
To address this gap, we introduce the \textit{Multimodal Conference Dataset (MCD)}, the first benchmark that integrates research papers, presentation videos, explanatory videos, and slides from the same works. We evaluate a range of embedding-based and vision–language models to assess their ability to discover fine-grained cross-format correspondences, establishing the first systematic benchmark for this task. Our results show that vision–language models are robust but struggle with fine-grained alignment, while embedding-based models capture text–visual correspondences well but equations and symbolic content form distinct clusters in the embedding space. These findings highlight both the strengths and limitations of current approaches and point to key directions for future research in multimodal scientific understanding. To ensure reproducibility, we release the resources for MCD at \href{https://github.com/meghamariamkm2002/MCD}{link}.
\vspace{-0.5cm}
\end{abstract}

%% file: sec/1_intro.tex
\section{Introduction}
\label{sec:intro}

In contemporary research, scientific communication extends far beyond traditional written papers. Discoveries and ideas are now shared through a rich ecosystem of materials—formal manuscripts, visual summaries, recorded presentations, and explanatory videos—each offering a distinct perspective on the same body of work. A written paper captures the complete technical depth, articulating methods, experiments, and analyses with precision. Visual summaries, such as slides highlight essential insights and illustrate complex ideas through concise design. Recorded presentations add the researcher’s voice, tone, and emphasis, revealing intent and interpretation that are often missing from written text. Explanatory videos, on the other hand, take a step toward accessibility, recontextualizing dense scientific content into forms that are easier to grasp and communicate to wider audiences.
Figure~\ref{fig:intro} shows these scientific communication formats—papers, slides, presentation videos, and explanation videos—and indicates their potential relationships using arrows.

Although these different forms describe the same research, they tend to exist in isolation. The links between them—how a figure in a paper connects to a slide image~\cite{10376585}, or how a spoken explanation corresponds to an equation—are rarely documented~\cite{10376585,10.1145/2467696.2467741,tp_pres,5705574,vid-slide-alignment,MaViLS}. This lack of correspondence creates information silos, where valuable insights conveyed through one medium remain disconnected from others. Consequently, searching for, comparing, or comprehensively understanding research across its various representations has become challenging. For students, this limits opportunities to learn from complementary materials; for researchers, it restricts automated analysis and knowledge discovery; and for systems that aim to organize or recommend research content, it reduces interpretability and coherence. 

Bridging these gaps requires structured alignment across different research representations.. Establishing meaningful connections between them can unlock new ways of exploring scientific knowledge, where a concept introduced in a paper can be directly linked to its visual explanation or a spoken commentary can guide a reader to the relevant figure or section. Such integration not only enhances accessibility and understanding but also supports the development of intelligent systems capable of reasoning across multiple forms of research communications. Motivated by this vision, we introduce a unified collection that brings together these diverse materials and offers a comprehensive view of how scientific ideas are conveyed, interpreted, and understood across formats.

We introduce the \textit{Multimodal Conference Dataset (MCD)}, the first benchmark for evaluating fine-grained correspondences across research papers, slides, presentation videos, and explanatory videos. Using MCD, we evaluate six models—embedding-based and vision--language—across three traversal settings: EV $\rightarrow$ PP(explanatory video to paper), S $\rightarrow$ PP(slide to paper), and PV $\rightarrow$ PP(presentation video to paper), covering paper segments: paragraphs, figures, equations, and algorithms. The results are analyzed across traversals, model types, and sizes. Our study provides several key insights. Vision--language models are robust across modalities, but their broad generality can make fine-grained alignment challenging. In embedding-based models, equations/symbolic content form distinct clusters in embedding space, showing limited mixing with text and visuals. Despite this, models achieve effective retrieval. GME-2.2B and InternVL3.5-38B are demonstrating solid performance in these segments. Through this comprehensive evaluation, we uncover trends and limitations across modalities, highlighting where models succeed, where they fail, and what challenges remain for robust cross-format understanding in scientific communication.

%% file: sec/2_relWork.tex
\section{Related Work}
\label{sec:relwork}

\subsection{Cross-Modal Retrieval}

Cross-modal retrieval aims to identify semantically corresponding information across different content types(text, images, audio, and video) enabling queries in one modality to retrieve relevant information in another. While widely studied in general domains such as image–text retrieval~\cite{frome2013devise,radford2021learning,alayrac2020self}, video–text alignment~\cite{miech2019howto100m,gabeur2020multi}, and vision–language understanding~\cite{li2021align,kim2021vilt}, these approaches typically focus on broad visual–linguistic content. Recent methods like E5‑V~\cite{DBLP:journals/corr/abs-2407-12580} adapt multimodal large language models (MLLMs) to produce universal embeddings across modalities via prompt-based representations. At the same time, GME~\cite{zhang2025gmeimprovinguniversalmultimodal} fine-tunes MLLM retrievers on large fused-modal datasets to support single-modal, cross-modal, and fused-modal retrieval.
In educational and research contexts, content is highly structured and semantically dense, combining precise text, abstract visuals, and spoken explanations. Early work aligned slides with papers~\cite{tp_pres,10.1145/2467696.2467741}, later enhanced with visual cues and sequential modeling~\cite{10.1145/2467696.2467741}. Research expanded to link presentation videos with slides~\cite{vid-slide-alignment,5705574,MaViLS}, integrating speech transcripts, visual frames, and temporal synchronization~\cite{multicross,10.1145/2647868.2654964}, as in the Google I/O dataset~\cite{10.1145/2647868.2654964}. Fine-grained figure–text associations were explored~\cite{10376585}, but prior work mostly considers pairwise correspondence. Our work addresses this gap by enabling comprehensive cross-modal retrieval across papers, slides, and videos through a unified benchmark.

\subsection{AI for Research}

The integration of Artificial Intelligence (AI) into the research lifecycle is transforming the creation, communication, and understanding of scientific knowledge. Traditionally, researchers manually prepared papers, slides, and presentations, but large language and multimodal models now enable automation and augmentation across these stages~\cite{doc2ppt,sun2025p2pautomatedpapertopostergeneration,zhu2025paper2videoautomaticvideogeneration,miao2025paper2agentreimaginingresearchpapers,shi2025presentagentmultimodalagentpresentation}. AI-driven systems assist in summarizing complex papers, generating multimodal representations such as slides~\cite{doc2ppt,AutoGenSlide,SlidesGen,mondal-etal-2024-presentations}, videos~\cite{shi2025presentagentmultimodalagentpresentation,zhu2025paper2videoautomaticvideogeneration}, and posters~\cite{sun2025p2pautomatedpapertopostergeneration}, and linking concepts across textual, visual, and spoken formats, enhancing efficiency and accessibility. Recent work has focused on automatic content generation from research papers. Slide generation models~\cite{doc2ppt,AutoGenSlide,SlidesGen} select key sections, figures, and equations to produce coherent decks, while video generation systems~\cite{zhu2025paper2videoautomaticvideogeneration,shi2025presentagentmultimodalagentpresentation} synthesize narrated, visual presentations. Poster frameworks~\cite{sun2025p2pautomatedpapertopostergeneration} provide concise, visually engaging summaries. Beyond static outputs, systems such as Paper2Agent~\cite{miao2025paper2agentreimaginingresearchpapers} transform papers into interactive AI agents that explain methods, reason about results, and engage in dialogue with users. Emerging methods also align spoken content in talks with corresponding regions in papers or slides, enabling highlighting of relevant sections during presentations~\cite{attendto2025}. This supports audiences in following explanations while navigating materials and maintaining context. Such alignment reduces cognitive load~\cite{Anmarkrud03042019,Cog2} and improves comprehension by directing attention to the most pertinent content. Together, these advances illustrate the role of AI in bridging modalities, reducing the effort required to prepare research artifacts, and enabling more interactive, multimodal, and accessible engagement with scientific knowledge. This convergence of AI and scholarly communication points to a dynamic, interpretable, and interconnected future for research dissemination.

%% file: sec/3_Dataset.tex
\section{Multimodal Conference Dataset}
\label{sec:MCD}
The \textit{Multimodal Conference Dataset (MCD)} is a curated collection of research papers, presentation slides, presentation videos, and explanatory videos that capture multiple perspectives of the same research work. It is designed to support the study of how related information is connected and expressed across these formats. Table~\ref{tab:comparison} presents a comparison of MCD with existing multimodal academic datasets.

\textbf{Data Collection:}
To construct MCD, we compiled materials that co-refer to the same research work. The dataset comprises two sets: the presentation set and an explanation set. The presentation set consists of paper–slide–presentation video triplets collected from the NeurIPS 2023 conference. Research papers were retrieved from \textit{arXiv}, while presentation slides and recorded talks were obtained from the \textit{SlidesLive} platform, which archives official conference presentations along with slide decks used by speakers.
The explanation set contains paper–explanatory video pairs, where explanatory videos were sourced from the \textit{Yannic Kilcher} YouTube channel\footnote{~\url{https://www.youtube.com/@YannicKilcher/videos}}. The corresponding papers were downloaded from the links provided in the video descriptions to ensure that the exact versions referenced in the explanations were used.

\begin{figure}[t]
  \centering
   \includegraphics[width=\linewidth]{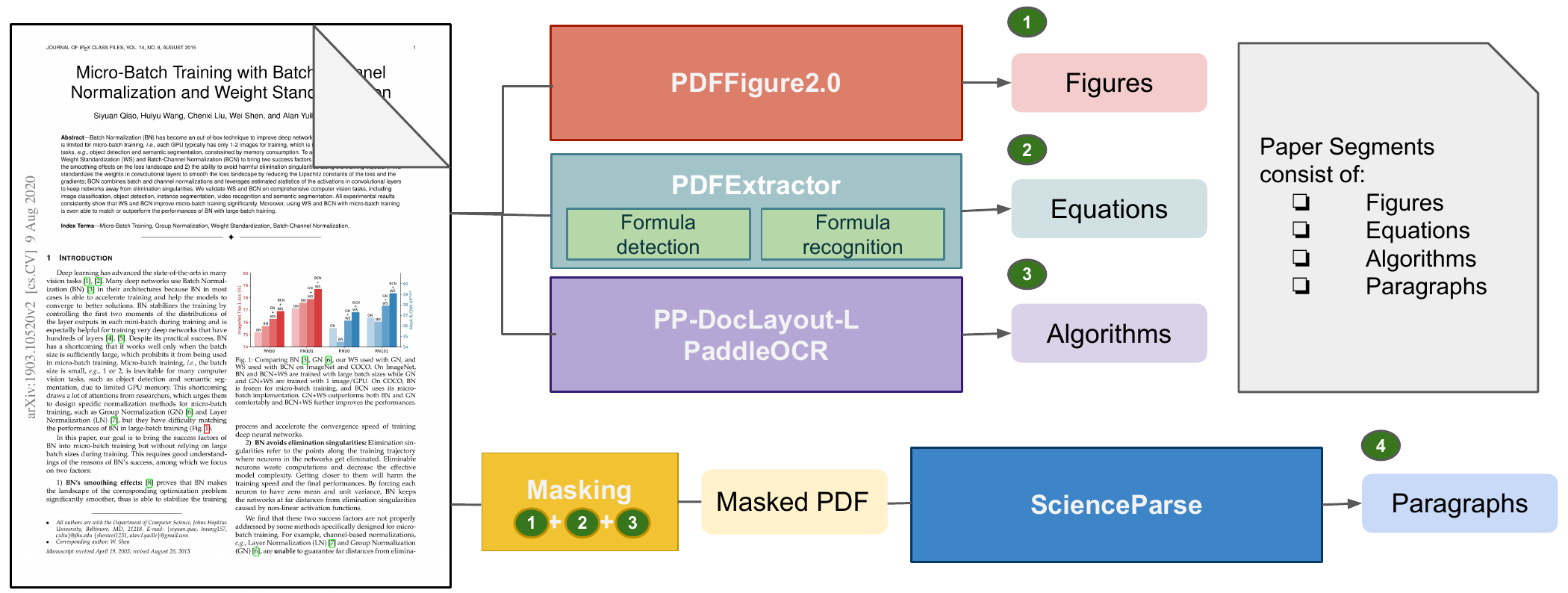}

   \caption{Pipeline for extracting paper segments—figures, equations, algorithms, and paragraphs—from the paper PDF.}
   \vspace{-0.6cm}
   \label{fig:paper_segments}
\end{figure}

\textbf{Data Preprocessing:}
For the presentation set, animated slides—where elements appeared incrementally across multiple slides-were identified, and only the final version containing all elements was retained to avoid redundancy. In such cases, the corresponding transcript segments up to the retained slides were concatenated to preserve the complete verbal context. Non-content slides, such as title, outline, acknowledgement, and closing (e.g., “Thank You”) slides were removed. After forming paragraph-level paper segments, segments containing only a single character were discarded~\ref{fig:paper_segments}. Additionally, segments categorized as “abandon” by \texttt{PDFExtractor} were filtered out from the paragraph set, and overlapping figure and equation elements were handled to ensure that only meaningful textual and visual components were preserved for subsequent alignment/retrieval tasks.

\begin{table}[t]
\centering
\scriptsize
\setlength{\tabcolsep}{2.5pt} 
\renewcommand{\arraystretch}{1.0} 
\begin{tabular}{lccccccc}
\toprule
\textbf{Dataset} & \multicolumn{5}{c}{\textbf{Contains}} & \textbf{Annot.} & \textbf{Task} \\ 
\cmidrule(lr){2-6}
 & \textbf{Sl.} & \textbf{Doc} & \textbf{PV} & \textbf{EV} & \textbf{Pos} & & \\ 
\midrule
DOC2PPT~\cite{doc2ppt} & \cmark & \cmark & \xmark & \xmark & \xmark & A & Slide Gen \\ 
Automatic slides generation~\cite{AutoGenSlide} & \cmark & \cmark & \xmark & \xmark & \xmark & M & Slide Gen \\ 
SciDuet~\cite{sun-etal-2021-d2s} & \cmark & \cmark & \xmark & \xmark & \xmark & A & Slide Gen \\ 
Persona-Aware D2S~\cite{mondal-etal-2024-presentations} & \cmark & \cmark & \xmark & \xmark & \xmark & A & Slide Gen \\ 
SlideAVSR~\cite{wang2024slideavsrdatasetpaperexplanation} & \xmark & \xmark & \cmark & \xmark & \xmark & A+M & AVSR \\ 
DocVideoQA~\cite{WangHG25-2} & \xmark & \xmark & \cmark & \xmark & \xmark & A+M & VideoQA \\ 
CS-PaperSum~\cite{DBLP:journals/corr/abs-2502-20582} & \xmark & \xmark & \cmark & \xmark & \xmark & A & Summ. \\ 
Paper2Poster~\cite{sun2025p2pautomatedpapertopostergeneration} & \xmark & \cmark & \xmark & \xmark & \cmark & A & Poster Gen \\ 
Doc2Present~\cite{shi2025presentagentmultimodalagentpresentation} & \cmark & \cmark & \cmark & \xmark & \xmark & A & Video Gen \\ 
Paper2Video~\cite{zhu2025paper2videoautomaticvideogeneration} & \cmark & \cmark & \cmark & \xmark & \xmark & A & Video Gen \\ 
\rowcolor{gray!15} 
\textbf{MCD} & \cmark & \cmark & \cmark & \cmark & \xmark & A+M & Cross Trav \\ 
\bottomrule
\end{tabular}
\vspace{-0.3cm}
\caption{Comparison of multimodal academic datasets based on included formats (Sl.: slides, Doc: documents, PV: presentation videos, EV: explanatory videos, Pos: posters), annotation type (A: automatic, M: manual), and supported task.}
\vspace{-0.4cm}
\label{tab:comparison}
\end{table}

\textbf{Data Statistics:}
The dataset comprises two primary subsets: the Presentation Set and the Explanation Set. The Presentation Set consists of 20 paper–slide–presentation video triplets, where each presentation video has an average duration of approximately 5 minutes. Segmentation is performed at the slide level, with each segment corresponding to an individual slide and its associated ASR transcript. In contrast, the Explanation Set includes 15 paper–explanatory video pairs, with explanatory videos averaging approximately 40 minutes in duration. For this set, segmentation follows predefined topic boundaries, and the provided segments are used directly to preserve coherent thematic units. Transcripts for all videos across both sets are generated using WhisperX. Table~\ref{tab:dataset_summary} summarizes the dataset, including the number of papers and slides, as well as the durations of Explanation Videos (EV) and Presentation Videos (PV). 

Figure~\ref{fig:stats} summarizes the dataset statistics for both subsets. Subfigures (a) and (d) show the distribution of content types (algorithms, equations, tables, and figures) in the papers. Subfigures (b) and (e) present frequent words from the sources (Explanation Set: ASR; Presentation Set: slide OCR + ASR): the Explanation Set features conversational terms (e.g., “see,” “right”), while the Presentation Set highlights technical terms (e.g., “model,” “data,” “algorithm”). Subfigures (c) and (f) depict segment counts per video, with the Explanation Set showing a more uniform distribution and the Presentation Set containing more videos with over ten segments.

\begin{table}[h!]
\centering
\vspace{-0.25cm}
\begin{tabular}{l|cccc}
\toprule
 & Count & Min & Max & Avg \\
\midrule
Ex. Videos (min.) & 15 & 25.55  & 72.36 & 39.77 \\
Slides & 20 & 4 & 19 & 10.60 \\
Pres. Videos (min.) & 20 & 2.25 & 5.33 & 4.58 \\
Papers (all) & 35 & 9 & 43 & 18.2 \\
\bottomrule
\end{tabular}
\vspace{-0.2cm}
\caption{Dataset content overview: counts of papers and slides, and durations (in minutes) for videos (EV = Explanation Video, PV = Presentation Video).}
\vspace{-0.3cm}
\label{tab:dataset_summary}
\end{table}

\textbf{Paper Segments:}
Each research paper comprises diverse elements such as text, figures, algorithms, and equations that collectively convey the study’s content. To systematically capture these components, we categorized the paper into four segment types: \textit{paragraphs}, \textit{equations}, \textit{figures} (including associated captions), and \textit{algorithms}.
Figures are extracted using \textsc{PDFFigures}~\cite{10.1145/2910896.2910904}. Algorithmic regions are detected with \textsc{PP-DocLayout-L}~\cite{sun2025ppdoclayoutunifieddocumentlayout}, from which bounding boxes are obtained and subsequently processed with \textsc{PaddleOCR}~\cite{DBLP:journals/corr/abs-2009-09941} to extract the text. Equations are identified and recognized using the \textsc{PDFExtractor} toolkit\footnote{\url{https://github.com/opendatalab/PDF-Extract-Kit?tab=readme-ov-file}}, specifically employing its formula detection and recognition modules.
A preprocessing stage ensures that extracted equations and figure regions are non-overlapping. These detected components are then removed from the PDF, and the remaining text is processed with \textsc{ScienceParse}\footnote{\url{https://reurl.cc/e62LXL}}, which segments the document into structured components such as \textit{abstract}, \textit{sections}, and \textit{authors}. The \textit{paragraph} segments were subsequently obtained from these textual components. Figure~\ref{fig:paper_segments} illustrates the overall extraction pipeline.

\textbf{Annotation:}
To establish fine-grained cross-modal correspondence, each source segment—whether a slide, a presentation video segment (slide + transcript), or an explanatory video segment (transcript)—is manually aligned with its corresponding paper segments. This ensures that all semantically relevant parts of the paper are accurately identified for every source segment. The dataset contains 15 explanatory videos and 20 presentation videos. The query set comprises 460 queries containing at least one paragraph, including 147 with at least one relevant figure, 121 with at least one relevant equation, and 56 with at least one relevant algorithm.

%% file: sec/4_probdefn.tex
\vspace{-0.2cm}
\section{Cross Linking Across Modalities}
\subsection{Cross Modal Grounding}
We address the task of linking multimodal research materials—explanatory videos, presentation videos, and slides—to their corresponding paper content. Each source provides a distinct perspective on the same work: the explanatory video emphasizes conceptual flow, the presentation video blends visuals and narration, and the slides convey condensed visual cues.

A research paper comprises multiple elements, including paragraphs, figures, equations, and algorithms. Here, a figure is defined as the visual image along with its associated captions (tables and their associated captions are also considered part of the figure). Given a source segment from any format—explanatory video, presentation video, or slide—the goal is to establish \textit{fine-grained, one-to-many correspondence} with the paper by identifying and ranking the most relevant elements. Formally, the task is:
\[
f: x \rightarrow \{p_i\}_{i=1}^{N},
\]
where $x$ denotes an input segment from a source modality and $\{p_i\}$ represents the set of paper elements semantically related to $x$. This formulation captures that a single segment may correspond to multiple paper components conveying the same underlying concept.

We consider three retrieval settings, each with a distinct query type. For explanatory videos, the query consists of the transcript of the video segment (EV$\rightarrow$PP). For slides, the query is the slide image (S$\rightarrow$PP). For presentation videos, the query combines both the slide image and the corresponding transcript (PV$\rightarrow$PP). Performance is assessed using \textit{NDCG@K} for paragraphs, figures, and equations, which measures how well the ranked paper elements align with human-annotated relevance judgments. For algorithms, we set a threshold of 0.6 and compute recall to evaluate retrieval.

%% file: sec/5_baselines.tex
\subsection{Are MLLMs ready?}
We evaluate six models covering embedding-based and vision–language models (VLMs). The embedding-based models include E5-V~\cite{DBLP:journals/corr/abs-2407-12580}, GME (2.2B, 8.2B)~\cite{zhang2025gmeimprovinguniversalmultimodal}, and ColQwen~\cite{faysse2024colpali}. E5-V produces universal embeddings by adapting multimodal large language models (MLLMs) with text-pair training and prompt-based bridging to align modalities, showing strong performance in image–text and document retrieval. GME supports single-modal, cross-modal, and fused-modal retrieval, with 2.2B and 8.2B variants (with and without instruction). Its large-scale instruction-based training on fused-modal datasets enables unified embeddings for text, images, and visual documents, making it effective for fine-grained correspondence between source segments and paper elements. ColQwen integrates multimodal information to produce unified representations for cross-modal retrieval.  

The VLMs—InternVL-4B, InternVL3.5-38B, and Qwen2.5-32B—are selected for their strong performance in OCR and document understanding. InternVL-4B is lightweight and efficient, InternVL3.5-38B provides high-quality multimodal comprehension, and Qwen2.5-32B offers strong visual--linguistic alignment. VLMs appear to understand individual media in tasks such as summarization, rephrasing, and question answering. We probe the depth of this understanding through the task of cross-linking and cross-grounding. Together, these embedding-based and VLM approaches enable evaluation across the three traversal tasks—EV$\to$PP, S$\to$PP, and PV$\to$PP—covering single-, cross-, and fused-modal retrieval. This setup ensures that each model’s strengths are leveraged to assess fine-grained multimodal correspondence in structured research content.

%% file: sec/7_res.tex
\section{Performance Analysis}
\begin{figure*}[t]
  \centering
   \includegraphics[width=\linewidth]{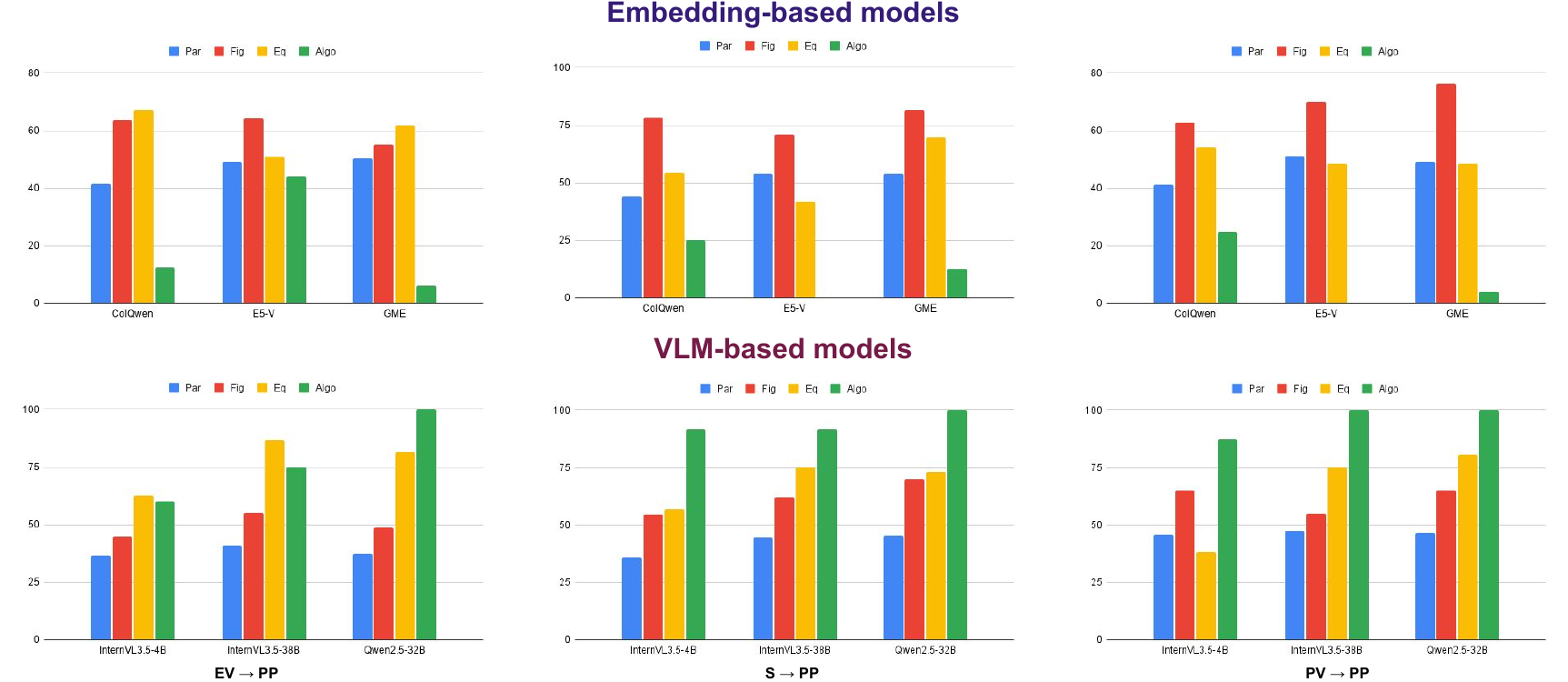}

   \caption{Illustrates the performance of embedding-based and VLM-based models across all three traversal settings (EV $\to$ PP, S $\to$ PP, PV $\to$ PP). NDCG@2 scores are used for paragraphs, figures, and equations. For algorithms, recall is calculated using a threshold of 0.6.}

   \label{fig:comp}
   \vspace{-0.5cm}
\end{figure*}

In this section, we analyze the retrieval performance of all models across the three traversal settings—explanatory video to paper (EV $\to$ PP), slides to paper (S $\to$ PP), and presentation video to paper (PV $\to$ PP)—and across different paper segment types, including paragraphs (par), figures (fig), equations (eq), and algorithms (algo) (Table~\ref{tab:results}). Our analysis is structured along four dimensions: (i) traversal-specific trends, (ii) embedding-based models, (iii) vision--language models, and (iv) the effect of model size, providing a comprehensive assessment of fine-grained multimodal alignment.
\subsection{Across Traversals}
Performance varies across the three traversals—EV$\to$PP, S$\to$PP, and PV$\to$PP—reflecting differences in multimodal correspondence (Table~\ref{tab:results}). The S$\to$PP traversal achieves the highest scores in majority cases due to strong visual–textual alignment between slides and paper content. Figures and equations on slides often directly reuse or paraphrase material from papers; however, algorithm retrieval remains challenging for most embedding-based models. In contrast, EV$\to$PP is the most difficult to traverse. Explanatory video transcripts contain rich narrative language that diverges from concise paper phrasing, reducing paragraph-level retrieval accuracy ($\leq 53$ NDCG@1 for the best embedding model). However, models such as GME-Qwen2VL-2B (with instructions) maintain stable figure-level performance, suggesting that conceptual anchors in narration aid alignment. PV$\to$PP combines the strengths of the other traversals. Slides provide visual grounding, and narration adds interpretive context. The performance is comparable to S$\to$PP and exceeds EV$\to$PP in the majority of cases. Overall, the trend S$\to$PP $>$ PV$\to$PP $>$ EV$\to$PP shows that retrieval correlates with semantic and structural continuity: clearer alignment supports stronger grounding, whereas stylistic divergence challenges fine-grained retrieval.

\begin{figure}[t]
  \centering
   \includegraphics[width=\linewidth]{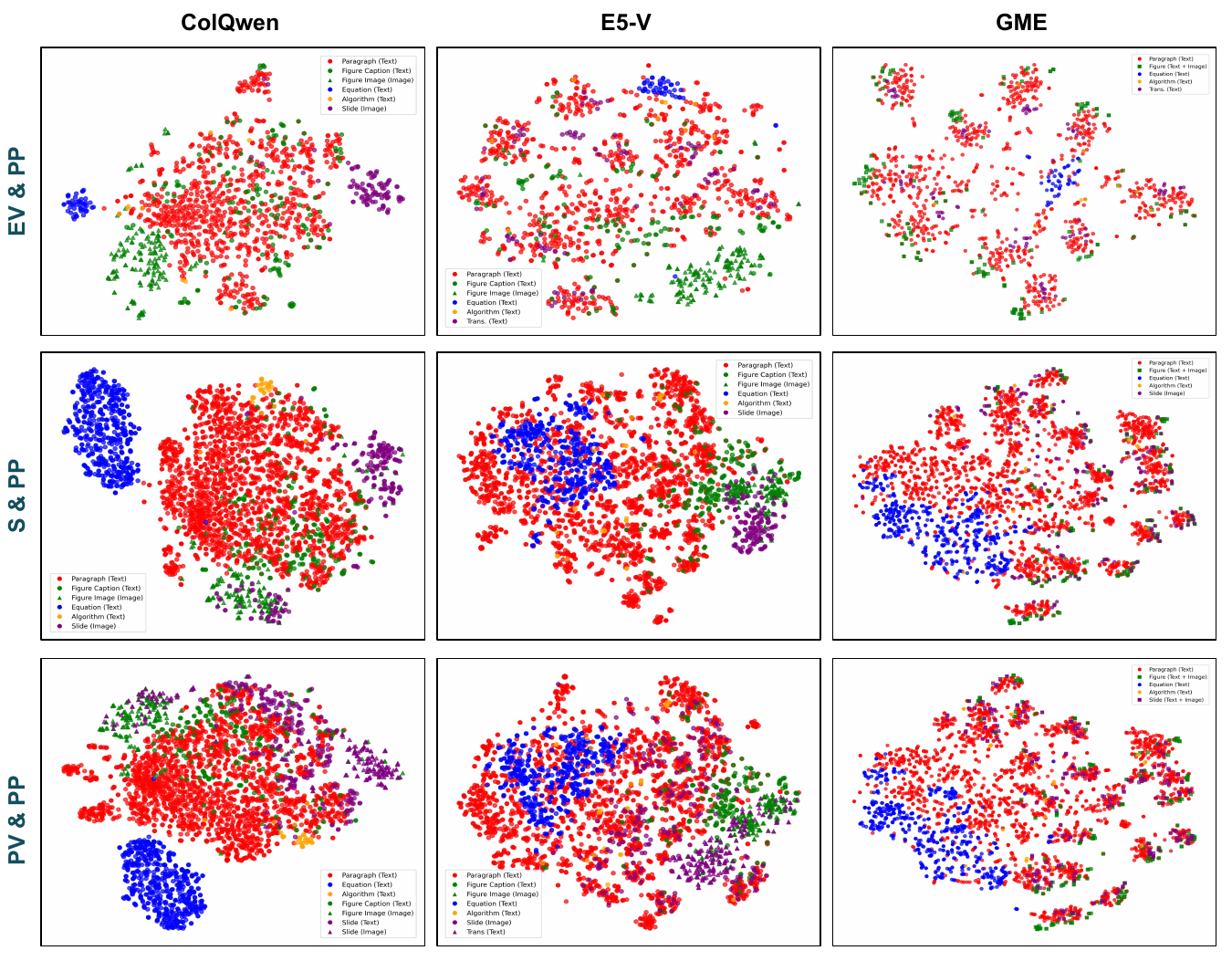}

   \caption{Visualization of query and candidate embeddings for representative instances across all three settings: EV $\to$ PP, S $\to$ PP, and PV $\to$ PP. Zoom in for a better view.}
   \label{fig:embed}
   \vspace{-0.3in}
\end{figure}

\begin{figure*}[t]
  \centering
   \includegraphics[width=\linewidth]{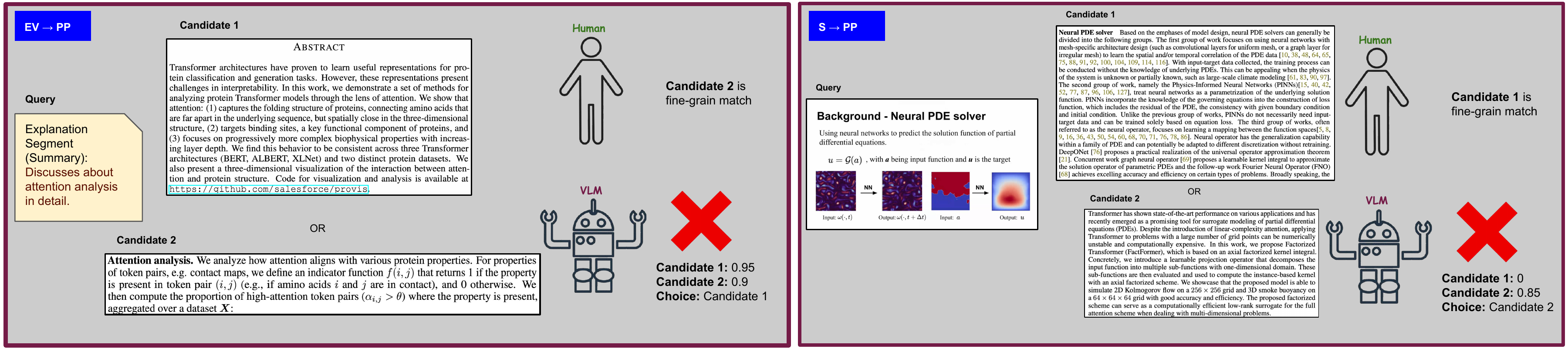}

   \caption{The figure presents two qualitative examples illustrating typical VLM failure cases. In the left example, the explanation segment discusses attention analysis in detail, as described for Candidate 2. However, the VLM assigns a higher similarity score to the abstract, which offers only a broad overview rather than the fine-grained correspondence expected. Similarly, in the right example, for the slide image, Candidate 1 provides a fine-grained match that closely aligns with the slide content, whereas Candidate 2 conveys only a general overview, yet receives a higher score from the VLM.}
   \label{fig:qual}
   \vspace{-0.5cm}
\end{figure*}

\subsection{Embedding-Based Models}

Figure~\ref{fig:embed} visualizes query embeddings—for explanatory videos (transcripts), slides (images), and presentation videos (slide + transcript)—alongside candidate embeddings of paragraphs, equations, figures, and algorithms across EV $\to$ PP, S $\to$ PP, and PV $\to$ PP traversals. Equation embeddings form compact, isolated clusters with limited mixing with textual and visual embeddings. In ColQwen, this separation is most pronounced, with equation clusters almost entirely distinct, indicating strong modality-specific encoding but weaker cross-modal integration. E5-V and GME show partial mixing: equation clusters remain identifiable but slightly overlap with paragraphs and figures, implying moderately shared semantic space.

Table~\ref{tab:results} reports quantitative performance. GME consistently performs strongly across traversals, with the 2.2B variants outperforming other embedding-based baselines for paragraphs, figures, and equations in most cases (but struggles with algorithms—unable to find any relevant matches). Across both the 2.2B and 7B variants, the use of instruction prompts does not lead to consistent improvements in performance. The GME-7B variant failed to retrieve relevant algorithms for S $\to$ PP. E5-V shows overall decent performance, but remains lower than GME in most cases. ColQwen2-VL has lower paragraph and figure scores but comparatively better equation retrieval in EV $\to$ PP and S $\to$ PP, with moderate performance in PV $\to$ PP. Notably, it achieves perfect scores for all algorithms across all traversals. Figure~\ref{fig:comp} (top row) presents NDCG@2 scores for all three models across paper segments.

\subsection{Vision Language Models}
We evaluate three vision–language models (VLMs)—two from the InternVL3.5 family (4B and 38B) and one from Qwen2.5-VL (32B)—along with proprietary Gemini models. Across most traversal settings, open-source VLMs exhibit lower performance all paragraph and figure retrieval in to embedding-based methods. This can be attributed to their inherently broad semantic sensitivity: they are robust at identifying subtle cross-modal relationships, often assigning high similarity scores to multiple partially relevant paper segments. While this reflects strong semantic understanding, it reduces precision of fine-grained segment-level correspondence, which is central to our task. In other words, although many candidate segments may be topically relevant, the task requires exact alignment, and VLMs’ generalization tendency leads to weaker discrimination among closely related candidates (see Figure~\ref{fig:embed}).

Interestingly, VLMs perform comparatively better in equation retrieval, likely due to the structured and symbolic nature of equations, which provide clearer alignment cues. The performance of the algorithms is consistently better than that of embedding-based models. All reported results for VLMs are based on few-shot inferences. Despite task-specific prompting and examples emphasizing fine-grained matching, these models tend to favor coarse semantic retrieval, excelling at capturing general relationships but lacking precision in detailed alignment.

In contrast, the closed-source Gemini models (Flash and Pro) consistently outperform open-source VLMs across all segment types and traversals. Gemini-2.5 Pro achieves the strongest overall results, with notable gains in paragraph and figure retrieval, while Flash offers competitive performance at a lower computational cost. Unlike open-source VLMs, Gemini models strike a better balance between semantic generalization and fine-grained matching, enabling more accurate retrieval.

Overall, the performance gap highlights a clear distinction: open-source VLMs are effective at capturing broad multimodal semantics but struggle with precise grounding, whereas closed-source models demonstrate stronger fine-grained alignments. Figure~\ref{fig:comp} (bottom row) shows the NDCG@2 scores for InternVL-38B, InternVL-4B, and Qwen-32B(Open-source VLMs) across different paper segments.

\begin{table*}[ht]
\centering  
\begin{tabular}{l|c|ccc|cc|cc|c}
\toprule
\multirow{2}{*}{\textbf{Model}} & \multirow{2}{*}{\textbf{Size}}
& \multicolumn{3}{c|}{\textbf{Par}} 
& \multicolumn{2}{c|}{\textbf{Fig}} 
& \multicolumn{2}{c|}{\textbf{Eq}} 
& \textbf{Algo} \\
\cmidrule(lr){3-5} \cmidrule(lr){6-7} \cmidrule(lr){8-9}
& & K=1 & K=2 & K=3 & K=1 & K=2 & K=1 & K=2 &  \\
\midrule
\multicolumn{10}{l}{\centering \textbf{Traversal: EV$\to$PP}} \\
\midrule
ColQwen2-VL~\cite{faysse2024colpali} & 2.21B & 47.27 & 41.52 & 40.37 & 61.40 & 63.80 & 63.16 & 67.01 & 12.50 \\
E5-V~\cite{DBLP:journals/corr/abs-2407-12580} & 8.4B & 50.29 & 49.05 & 47.50 & 62 & 64.40 & 39.71 & 50.84 & 44 \\
GME-Qwen2VL-2B w/o instr.~\cite{zhang2025gmeimprovinguniversalmultimodal} & 2.2B & 53.64 & 48.10 & 46.28 & 54.39 & 56.96 & 57.89 & 61.75 & 6.25 \\
GME-Qwen2VL-2B with instr.~\cite{zhang2025gmeimprovinguniversalmultimodal} & 2.2B & 53.64 & 50.21 & 49.70 & 50.88 & 55.23 & 52.63 & 61.84 & 6.25  \\
GME-Qwen2VL-7B w/o instr.~\cite{zhang2025gmeimprovinguniversalmultimodal} & 8.2B & 47.27 & 44.90 & 45.18 & 38.60 & 45.67 & 47.37 & 47.90 & 62.5 \\
GME-Qwen2VL-7B with instr.~\cite{zhang2025gmeimprovinguniversalmultimodal} & 8.2B & 50.91 & 48.19 & 46.47 & 47.37 & 51.30 & 47.37 & 53.26 & 62.5 \\
InternVL3.5-4B~\cite{wang2025internvl35advancingopensourcemultimodal} & 4B & 36.29 & 36.45 & 34.79 & 39.34 & 44.68 & 61.90  & 62.39 & 60 \\
InternVL3.5-38B~\cite{wang2025internvl35advancingopensourcemultimodal} & 38B & 39.09 & 40.72 & 40.71 & 47.37 & 55.12 & 84.21 & 86.78 & 75 \\
Qwen2.5-VL-32B~\cite{qwen2025qwen25technicalreport} & 32B & 34.55 & 37.19 & 38.98 & 43.86 & 48.72 & 73.68 & 81.61 & 100 \\

Gemini-2.5 Flash & - & 53.64 & 53.60 & 53.72 & 52.63 & 56.38  &  73.68 & 82.36 &  75\\
Gemini-2.5 Pro & - & 62.73 & 60.84 & 60.28 & 71.93 & 77.89 & 94.74 & 88.63 & 75 \\
\midrule

\multicolumn{10}{l}{\centering \textbf{Traversal: S$\to$PP}} \\
\midrule
ColQwen2-VL~\cite{faysse2024colpali} & 2.21B & 47.88 & 44.02 & 45.03 & 68.18 & 78.22 & 50.98 & 54.21 & 25 \\
E5-V~\cite{DBLP:journals/corr/abs-2407-12580} & 8.4B & 56.97 & 54.02 & 54.80 & 63.64 & 70.81 & 43.14 & 41.82 & 0 \\
GME-Qwen2VL-2B w/o instr.~\cite{zhang2025gmeimprovinguniversalmultimodal} & 2.2B & 58.79 & 53.58 & 54.48 & 75 & 83.60 & 68.63 & 70.54 & 8.33 \\
GME-Qwen2VL-2B with instr.~\cite{zhang2025gmeimprovinguniversalmultimodal} & 2.2B & 58.18 & 54.08 & 54.76 & 72.73 & 81.33 & 68.63 & 69.78 & 12.50 \\
GME-Qwen2VL-7B w/o instr.~\cite{zhang2025gmeimprovinguniversalmultimodal} & 8.2B & 52.12 & 50 & 51.45 & 77.27 & 83.01 & 66.67 & 66.39 & 0 \\
GME-Qwen2VL-7B with instr.~\cite{zhang2025gmeimprovinguniversalmultimodal} & 8.2B & 55.15 & 52.35 & 53.89 & 77.27 & 85.88 &  68.63 & 68.35  & 0 \\
InternVL3.5-4B~\cite{wang2025internvl35advancingopensourcemultimodal} & 4B & 33.33 & 35.92 & 37.71 & 40.91 & 54.69 & 50.98 & 56.97  &  91.67 \\
InternVL3.5-38B~\cite{wang2025internvl35advancingopensourcemultimodal} & 38B & 42.42 & 44.58 & 47.55 & 50.00 & 62.03 & 70.59 & 75.06 & 91.67 \\
Qwen2.5-VL-32B~\cite{qwen2025qwen25technicalreport} & 32B & 46.06 & 45.60 & 50.60 & 59.09 & 70.01 & 64.71 & 73.17  & 100 \\

Gemini-2.5 Flash & - & 59.39 & 57.91 & 60.68 & 77.27 & 86.43 & 80.39 & 87.81 &  100\\
Gemini-2.5 Pro & - & 66.06 & 62.85  & 64.87 & 79.55 & 85.84 & 90.20 & 88.40 & 100  \\

\midrule

\multicolumn{10}{l}{\centering \textbf{Traversal: PV$\to$PP}} \\
\midrule
ColQwen2-VL~\cite{faysse2024colpali} & 2.21B & 46.49 & 41.16 & 41.90 & 47.83 & 62.91 & 50.98 & 54.21 & 25 \\
E5-V~\cite{DBLP:journals/corr/abs-2407-12580} & 8.4B & 54.05 & 51.29 & 53.91 & 60.87 & 69.94 & 47.06 & 48.49 & 0 \\
GME-Qwen2VL-2B w/o instr.~\cite{zhang2025gmeimprovinguniversalmultimodal} & 2.2B & 52.97 & 48.06 & 48.57 & 65.22 & 76.72 & 49.02 & 49.70 & 4.17 \\
GME-Qwen2VL-2B with instr.~\cite{zhang2025gmeimprovinguniversalmultimodal} & 2.2B & 55.14 & 49.24 & 49.85 & 67.39 & 76.15 & 45.10 & 48.53 & 4.17 \\
GME-Qwen2VL-7B w/o instr.~\cite{zhang2025gmeimprovinguniversalmultimodal} & 8.2B & 55.14 & 49.39 & 49.63 & 69.57 & 79.70 & 60.78 & 61.26 & 0 \\
GME-Qwen2VL-7B with instr.~\cite{zhang2025gmeimprovinguniversalmultimodal} & 8.2B & 55.68 & 49.30 & 50.02 & 65.22 & 78.40 & 62.75 & 62.47 & 4.17 \\
InternVL3.5-4B~\cite{wang2025internvl35advancingopensourcemultimodal} & 4B & 46.56 & 45.88 & 47.32 & 59.18 & 65.12 & 37.25 & 38.21 & 87.50 \\
InternVL3.5-38B~\cite{wang2025internvl35advancingopensourcemultimodal} & 38B & 49.73 & 47.70 & 48.98 & 47.83 & 55.21 & 75.00 & 75.00 & 100 \\
Qwen2.5-VL-32B~\cite{qwen2025qwen25technicalreport} & 32B & 44.86 & 46.39 & 49.41 & 52.17 & 65.05 & 74.51 & 80.70 &  100\\

Gemini-2.5 Pro & - & 59.46 & 58.66 & 61.20 & 78.26 & 81.53 & 86.27  & 89.99 & 100 \\
Gemini-2.5 Flash & - & 59.46 & 56.22 & 57.22 & 76.09 & 81.57 & 82.35 & 86.06 &  100\\
\bottomrule
\end{tabular}

\caption{Retrieval results for traversals from explanatory video (EV$\rightarrow$PP), slides (S$\rightarrow$PP), and presentation video (PV$\rightarrow$PP) to paper. Evaluation is performed over paper candidates—including paragraphs (Par), figures (Fig), and equations (Eq)—using NDCG@K, while for algorithms (Algo), only candidates with a threshold greater than 0.6 are considered, using recall as the metric. All values are reported in percentage (\%).}
\vspace{-0.6cm}
\label{tab:results}
\end{table*}

\vspace{-0.1cm}
\subsection{Effect of Model Size}
Within the GME variants 2.2B and 8.2B, an increase in model size does not consistently translate to improved performance. The 2.2B variant often matches or even surpasses the 8.2B model across multiple traversals, indicating that scaling alone does not guarantee better multimodal alignments. No clear trend correlating the model size with the retrieval quality is observed among the embedding-based methods. Similarly, larger embedding-based models such as E5-V (8.4B) and GME-8.2B do not exhibit substantial gains over smaller counterparts like GME-2.2B or ColQwen2-VL (2.2B).
In contrast, among the VLM-based models, increasing model size leads to significant and consistent improvements, particularly within the InternVL family. The performance rise is especially pronounced for equation retrieval, where the larger models demonstrate a substantial advantage across all three traversal settings  (EV$\to$PP, S$\to$PP, and PV$\to$PP), highlighting the benefits of scaling in vision–language architectures. 
\newline
Overall, these observations indicate that model performance depends on a combination of traversal type, segment modality, and model architecture. Embedding-based models tend to struggle with algorithms, while VLMs struggle with paragraphs; however, VLMs show notable gains from scaling, particularly for equations. This underscores that effective fine-grained retrieval relies on both modality alignment and model design, shaping performance across papers, slides, and videos.

%% file: sec/8_conclusion.tex
\vspace{-0.2cm}
\section{Conclusion}
\vspace{-0.15cm}
We introduce the \textit{Multimodal Conference Dataset (MCD)}, the first benchmark for fine-grained correspondences across research papers, slides, presentation videos, and explanatory videos. MCD evaluates six embedding-based and vision–language models across three traversals—EV $\rightarrow$ PP, S $\rightarrow$ PP, and PV $\rightarrow$ PP. Vision–language models are robust but less precise for fine-grained alignment, while embedding-based models capture text–visual correspondences; equations and symbolic content form distinct clusters, showing limited mixing yet remaining retrievable. GME-2.2B and InternVL3.5-38B perform well. Overall, MCD provides a unified framework for benchmarking cross-format scientific retrieval, highlighting model strengths and limitations and guiding future research in fine-grained correspondence discovery.

%% file: suppl.tex
\FloatBarrier
\appendix
\section*{Appendix}

\section{Benchmark Utility}
The core contribution of this benchmark is to enable seamless navigation and alignment across different forms of scientific communication—such as research papers, presentation slides, recorded talks, and explanatory videos—each of which presents information with a distinct structure, level of detail, and perspective. While all these media aim to convey the same underlying concepts, they do so in complementary ways: papers provide formal rigor and completeness, slides emphasize key ideas and visual summaries, presentation videos capture the narrative flow and intent of the speaker, and explanation videos often simplify and reinterpret concepts for clarity.

This diversity, while valuable, creates a practical challenge for learners and researchers: connecting corresponding pieces of information across modalities is non-trivial. A concept introduced in slides or a presentation video may not map to a single element, but instead correspond to multiple components within a paper—for example, a detailed paragraph, a formal equation, and an associated figure. Similarly, content from explanatory videos may align with complementary parts of the paper that provide intuition, derivation, or formal grounding. Identifying these one-to-many correspondences requires both semantic understanding and fine-grained alignment across modalities.

Additionally, by enabling such fine-grained alignment, the benchmark can also support the evaluation of paper-to-slide and paper-to-video generation systems, as well as the probing of multimodal LLM grounding in scientific domains. It is not intended for end users today, but rather for evaluating systems that will power such tools.
\begin{figure*}[b]
\centering
\includegraphics[width=\linewidth]{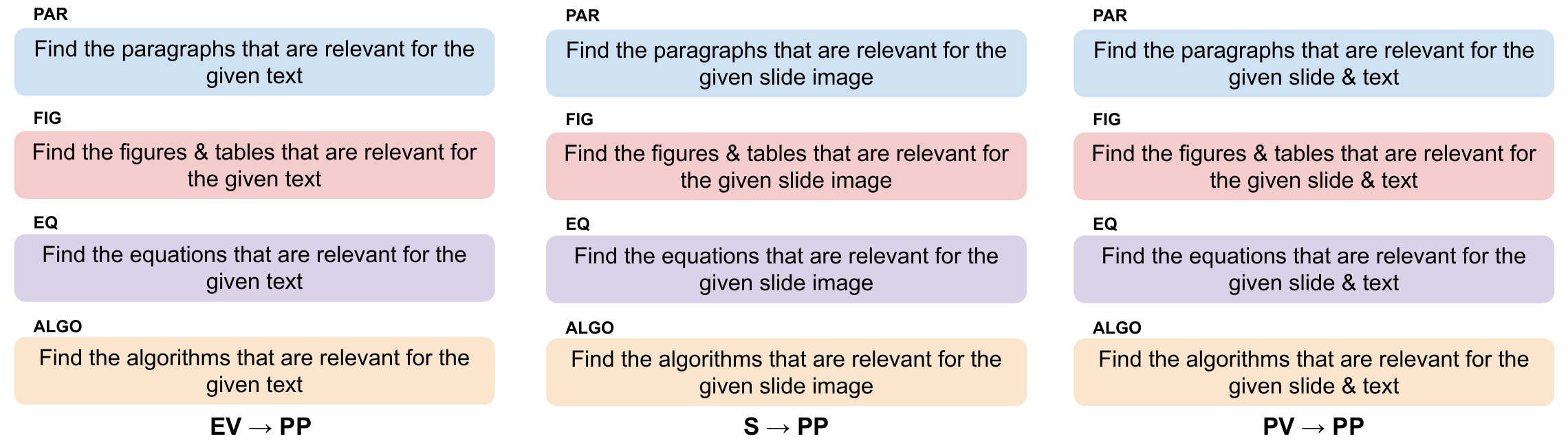}
\caption{The figure illustrates the instruction prompts used for the GME models across the three traversal settings: EV~$\rightarrow$~PP, S~$\rightarrow$~PP, and PV~$\rightarrow$~PP.}
\label{fig:gme_instr}
\end{figure*}

\vspace{0.8cm}
\section{Dataset Details}
 The dataset consists of 15 explanatory videos and 20 presentation videos. It includes 142 explanatory video segments, of which 128 have at least one corresponding relevant paper segment. Additionally, there are 204 slides and 204 presentation video segments, of which 200 have at least one relevant paper segment.
The query set comprises 460 queries, each containing at least one paragraph. Among these, 147 queries include at least one relevant figure, 121 include at least one relevant equation, and 56 include at least one relevant algorithm.

\subsection{Preprocessing and Data Quality}
\paragraph{ASR Transcripts.}
All video transcripts are generated using \textit{WhisperX}, ensuring consistent and high-quality automatic speech recognition across the dataset. To validate transcription reliability, we performed manual spot-checks on a sampled subset of videos, observing an accuracy of greater than $90\%$.
\paragraph{Slide Content}
We do not apply explicit OCR to slide images. Instead, slide-based modalities rely directly on raw visual inputs. In the \emph{S$\rightarrow$PP} setting, models are provided with full slide images, requiring them to jointly interpret visual structure and embedded text. In the \emph{PV$\rightarrow$PP} setting, slide images are paired with ASR transcripts, enabling complementary use of visual and spoken information.

\paragraph{Timestamp Alignment.}
Temporal alignment between video and corresponding content is derived from reliable, source-specific signals rather than post hoc synchronization. For \emph{EV$\rightarrow$PP}, we use explicit topic timestamps provided in YouTube videos, which segment content into semantically meaningful units. For \emph{PV$\rightarrow$PP}, we leverage the native alignment between slides and speech available in SlidesLive recordings.

\subsection{Annotation Details} 
A pilot study with four annotators on a sampled MCD subset showed complete agreement, indicating that the task is largely unambiguous in the scientific literature domain. This high level of agreement is likely due to the structured and precise nature of scientific content. Based on this observation, the remainder of the dataset was annotated by a single high-quality expert to ensure consistency and efficiency.

While the segment types are clearly defined, special care was taken to standardize the criteria for determining correct correspondences, particularly in cases that may appear ambiguous or partial. In general, annotations are guided by conceptual relevance rather than strict surface-level matching. For instance, figures in slides are considered aligned with those in papers even when they are cropped, reformatted, or stylistically modified, provided they convey the same underlying concept; in such cases, both visual content and accompanying captions are taken into account. Similarly, equation alignment does not require exact symbolic equivalence, but instead focuses on whether the expressions represent the same concept or play an equivalent role in the explanation. These guidelines facilitate systematic and consistent annotation.

\section{Baseline Details}
In this section, we discuss the details regarding the baseline models, both vision language models and embedding-based models. 
 
 All models are evaluated in their default, off-the-shelf settings without any fine-tuning or parameter modification. This includes using each model with its standard resolution, context length, and inference configurations as provided by their respective implementations. Closed-source models such as Gemini are also used under their default API settings
 
\subsection{Embedding-based models}
For the embedding-based baselines, ColQwen, E5-V, and four configurations of GME are evaluated, corresponding to two model sizes (2.2B and 8.2B), each used with and without instructions. Figure~\ref{fig:gme_instr} shows the instruction prompts used for the GME variants.

For ColQwen and E5-V, when a candidate or query contains fused content (e.g., an image and its caption), similarity scores are computed independently for each component and the maximum is taken. In contrast, GME uses a single fused embedding for such multimodal inputs.

All images are provided to the respective model processors in their original resolution, without any manual resizing or preprocessing, allowing each model to internally handle image scaling according to its default pipeline. Image resolution can influence embedding quality and retrieval performance; using the native preprocessing of each model ensures consistency in evaluation and avoids introducing model-specific biases through external adjustments.
\subsection{Vision language models}
\begin{figure*}[b]
\centering
\includegraphics[height=0.95\textheight,width=0.9\linewidth]{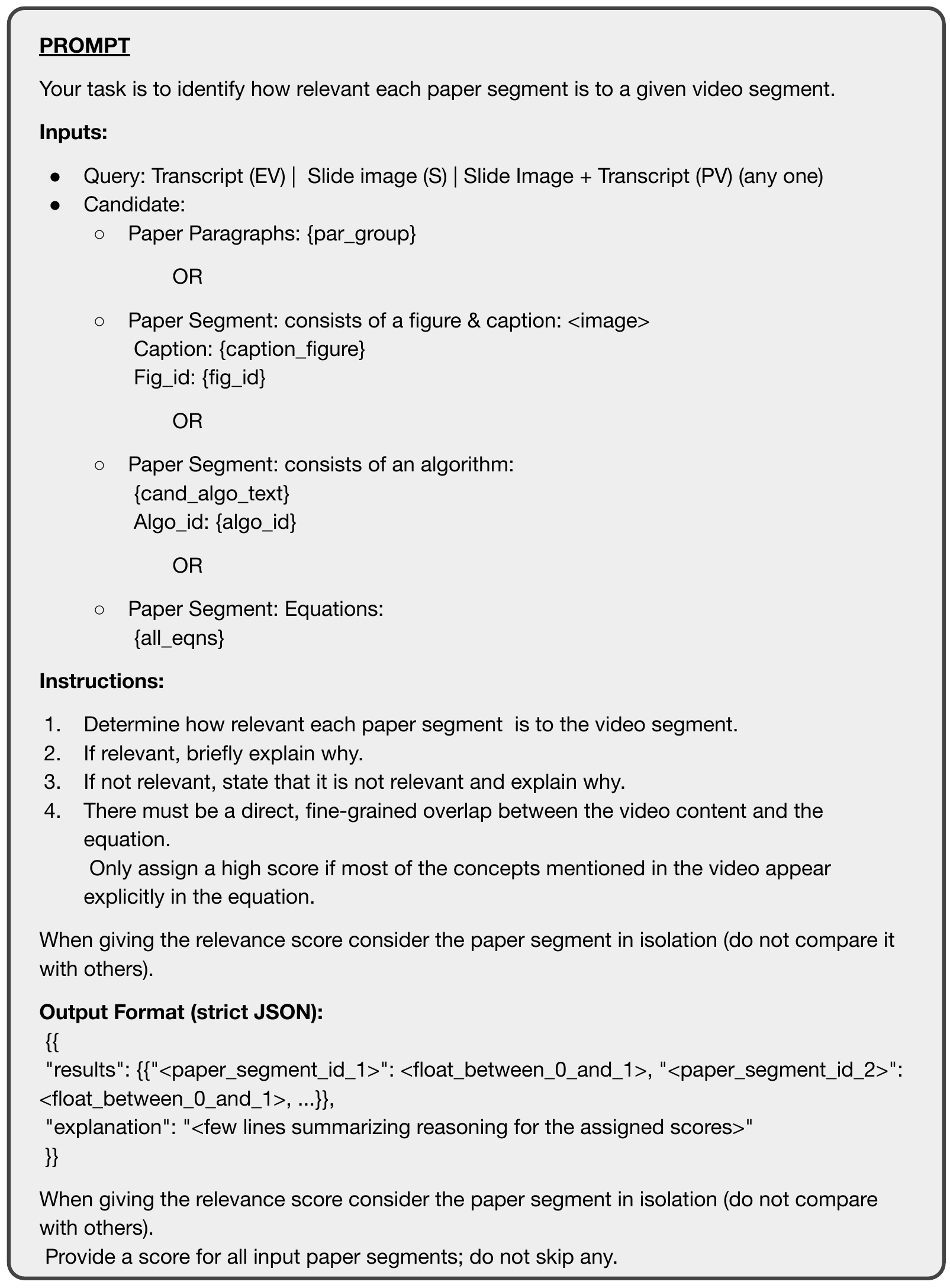}
\caption{Prompt used for VLM evaluation. Queries are provided as: transcript (explanatory video), slide (slides), or slide+transcript (presentation videos). Each query is evaluated against four paper segment types: paragraph, figure, equation, and algorithm.}
\label{fig:vlm_prompt}
\end{figure*}

To encourage fine-grained correspondence, we include keywords such as “direct” and “fine-grained” in the VLM prompts. We use all vision-language models in their default settings, without any additional fine-tuning or parameter modification. Because supplying all paper segments (e.g., paragraphs) often exceeds the model’s maximum context length—and evaluating segments one by one is computationally expensive—we provide the model with a subset of segments at a time. The VLM is instructed to evaluate each segment in isolation and assign a relevance score. These scores are then sorted in descending order to compute rank-based retrieval metrics. Figure~\ref{fig:vlm_prompt} shows the prompt used.

Since slide images are provided without explicit OCR, the retrieval performance of VLMs inherently depends on their implicit ability to read and interpret textual content within images. To ensure that the task remains well-posed and does not disproportionately disadvantage certain models, we conducted preliminary checks verifying that the evaluated VLMs can reliably extract and reproduce the textual content from slides. This ensures that the task primarily evaluates cross-modal alignment rather than raw text recognition ability.

\section{Evaluation Details}
For paragraph, figure, and equation retrieval, we report NDCG@K to evaluate ranking quality. Algorithm retrieval is treated separately, as most papers contain at most one algorithm, making NDCG@1 uninformative (i.e., trivially 100 when the correct item is retrieved).

We instead evaluate algorithm retrieval using recall at a similarity threshold. Specifically, we sweep thresholds from $0.3$ to $0.95$ in increments of $0.05$ for \emph{EV$\rightarrow$PP}, \emph{S$\rightarrow$PP}, and \emph{PV$\rightarrow$PP}, and select $0.6$ as a stable operating point. Lower thresholds tend to introduce false positives, while higher thresholds are overly strict and reduce recall. Precision is not reported, as most queries have only one relevant algorithm(not all), making recall the more informative metric. There are cases where queries have a single relevant algorithm, making retrieval effectively binary, which is one of the reasons we observe extreme values (0 or 100) for algorithms.

\subsection{Annotation Granularity}
Paper segments are annotated at four levels: paragraphs (text), figures (image + caption, including tables), equations (text), and algorithms (text). Queries are defined as ASR transcripts for explanatory videos, slide images for slides, and both for presentation videos.
. 

\section{Ablation}
\subsection{0-Shot and k-Shot Performance Analysis}
From Table~\ref{tab:ablation}, we observe that, in most cases, the k-shot setting outperforms the zero-shot setting. Providing a small number of input–output examples helps vision–language models establish finer-grained correspondences. Notably, improvements are consistent across paragraph, figure, and equation retrieval. In contrast, the performance gains for algorithm retrieval are negligible, suggesting that example-based guidance offers limited additional benefit for this modality.
\section{Additional Dataset Details}

\begin{table}[h]
\centering
\begin{tabular}{lccc}
\toprule
\textbf{Modality} & \textbf{Avg (\#)} & \textbf{Max (\#)} & \textbf{Min (\#)} \\
\midrule
Slides (S)              & 108.17 & 476 & 6 \\
Slide + Transcript (PV) & 156.8 & 582 & 39 \\
Transcript (EV)         & 615.84  & 2759 & 28 \\
\bottomrule
\end{tabular}
\caption{Word count statistics across the different content types(EV, PV, S), showing average, maximum, and minimum counts.}
\label{tab:wordcount}
\end{table}

\begin{table}[h]
\centering
\begin{tabular}{lcc}
\toprule
\textbf{Category} & \textbf{Presentation Set} & \textbf{Explanation Set} \\
\midrule
Algorithm          & 30 & 19 \\
Equation           & 579 & 135 \\
Image + Caption    & 107 & 103 \\
Table + Caption    & 88  & 82     \\
\bottomrule
\end{tabular}
\caption{Distribution of paper segments—algorithms, equations, images, and tables—across the Presentation and Explanation Sets.}
\label{tab:content_stats}
\end{table}

\begin{table*}[ht]
\centering
\begin{tabular}{l|c|ccc|cc|cc|c}
\toprule
\multirow{2}{*}{\textbf{Model}} & \multirow{2}{*}{\textbf{Size}}
& \multicolumn{3}{c|}{\textbf{Par}} 
& \multicolumn{2}{c|}{\textbf{Fig}} 
& \multicolumn{2}{c|}{\textbf{Eq}} 
& \textbf{Algo} \\
\cmidrule(lr){3-5} \cmidrule(lr){6-7} \cmidrule(lr){8-9}
& & K=1 & K=2 & K=3 & K=1 & K=2 & K=1 & K=2 &  \\
\midrule
\multicolumn{10}{l}{\centering \textbf{Traversal: EV$\to$PP}} \\
\midrule
InternVL3.5-4B (0-shot)~\cite{wang2025internvl35advancingopensourcemultimodal} & 4B & 29.87 &32.51 & 32.30 & 25.58 & 33.15 & 50.00 & 47.06 & 0\\
InternVL3.5-4B (k-shot)~\cite{wang2025internvl35advancingopensourcemultimodal} & 4B & 36.29 & 36.45 & 34.79 & 39.34 & 44.68 & 61.90  & 62.39 & 60 \\
InternVL3.5-38B (0-shot)~\cite{wang2025internvl35advancingopensourcemultimodal} & 38B & 35.06 & 32.37 & 35.93 & 34.88 & 41.89 & 61.11 & 70.27 & 100 \\
InternVL3.5-38B (k-shot)~\cite{wang2025internvl35advancingopensourcemultimodal} & 38B & 39.09 & 40.72 & 40.71 & 47.37 & 55.12 & 84.21 & 86.78 & 75 \\
Qwen2.5-VL-32B (0-shot)~\cite{qwen2025qwen25technicalreport} & 32B & 25.00 & 32.51 & 34.66 & 18.60 & 35.08 & 66.67 &  75.03 & 100 \\

Qwen2.5-VL-32B (k-shot)~\cite{qwen2025qwen25technicalreport} & 32B & 34.55 & 37.19 & 38.98 & 43.86 & 48.72 & 73.68 & 81.61 & 100 \\
\midrule

\multicolumn{10}{l}{\centering \textbf{Traversal: S$\to$PP}} \\
\midrule
InternVL3.5-4B (0-shot) ~\cite{wang2025internvl35advancingopensourcemultimodal} & 4B & 30.30 &33.83 & 35.50 & 40.91 & 54.69 &    50.98 & 53.45 & 90.48\\
InternVL3.5-4B (k-shot)~\cite{wang2025internvl35advancingopensourcemultimodal} & 4B & 33.33 & 35.92 & 37.71 & 40.91 & 54.69 & 50.98 & 56.97  &  91.67 \\
InternVL3.5-38B (0-shot)~\cite{wang2025internvl35advancingopensourcemultimodal} & 38B & 42.31 & 43.44 & 44.75 & 42.86 & 51.58 & 65.52 & 72.29 & 91.67  \\
InternVL3.5-38B (k-shot)~\cite{wang2025internvl35advancingopensourcemultimodal} & 38B & 42.42 & 44.58 & 47.55 & 50.00 & 62.03 & 70.59 & 75.06 & 91.67 \\
Qwen2.5-VL-32B (0-shot)~\cite{qwen2025qwen25technicalreport} & 32B & 43.03 & 45.19 & 50.82  & 50.00 & 66.10 & 62.75 & 73.20 & 91.67 \\

Qwen2.5-VL-32B (k-shot)~\cite{qwen2025qwen25technicalreport} & 32B & 46.06 & 45.60 & 50.60 & 59.09 & 70.01 & 64.71 & 73.17  & 100 \\
\midrule

\multicolumn{10}{l}{\centering \textbf{Traversal: PV$\to$PP}} \\
\midrule
InternVL3.5-4B (0-shot)~\cite{wang2025internvl35advancingopensourcemultimodal} & 4B & 38.92 & 38.11 & 40.09 & 47.83 & 61.01 & 31.37 & 36.80 & 87.50  \\
InternVL3.5-4B (k-shot)~\cite{wang2025internvl35advancingopensourcemultimodal} & 4B & 46.56 & 45.88 & 47.32 & 59.18 & 65.12 & 37.25 & 38.21 & 87.50 \\
InternVL3.5-38B (0-shot)~\cite{wang2025internvl35advancingopensourcemultimodal} & 38B & 48.11 & 47.04 & 49.21 & 43.48 & 54.98 & 75.51 & 75.51 & 100 \\
InternVL3.5-38B (k-shot)~\cite{wang2025internvl35advancingopensourcemultimodal} & 38B & 49.73 & 47.70 & 48.98 & 47.83 & 55.21 & 75.00 & 75.00 & 100 \\
Qwen2.5-VL-32B (0-shot)~\cite{qwen2025qwen25technicalreport} & 32B & 40.54 & 45.32 & 47.84 & 52.17 & 65.00 & 74.50 & 80.00  & 95.65\\
Qwen2.5-VL-32B (k-shot)~\cite{qwen2025qwen25technicalreport} & 32B & 44.86 & 46.39 & 49.41 & 52.17 & 65.05 & 74.51 & 80.70 &  100\\
\bottomrule
\end{tabular}
\caption{
Zero-shot and k-shot retrieval performance for traversals from explanatory video (EV$\rightarrow$PP), slides (S$\rightarrow$PP), and presentation video (PV$\rightarrow$PP) to paper content. Models are evaluated over paragraph (Par), figure (Fig), and equation (Eq) candidates using NDCG@K, while algorithm (Algo) retrieval is evaluated using recall over candidates exceeding a similarity threshold of 0.6. Zero-shot rows contain no in-context examples, whereas one-shot rows incorporate a single in-context demonstration. All values are reported in percentage (\%).}
\vspace{-0.5cm}
\label{tab:ablation}
\end{table*}

%% file: main.bib
@String(CVPR= {IEEE Conf. Comput. Vis. Pattern Recog.})

@String(ICCV= {Int. Conf. Comput. Vis.})

@String(ECCV= {Eur. Conf. Comput. Vis.})

@String(ICPR = {Int. Conf. Pattern Recog.})

@String(ACMMM= {ACM Int. Conf. Multimedia})

@String(ICASSP=	{ICASSP})

@String(ICLR = {Int. Conf. Learn. Represent.})

@String(AAAI = {AAAI})

@String(CVPR  = {CVPR})

@String(ICCV  = {ICCV})

@String(ECCV  = {ECCV})

@String(ICPR  = {ICPR})

@String(ACMMM = {ACM MM})

@String(ICLR  = {ICLR})

@article{5705574,
author={Fan, Quanfu and Barnard, Kobus and Amir, Arnon and Efrat, Alon},
title = {Robust Spatiotemporal Matching of Electronic Slides to Presentation Videos},
journal={IEEE Transactions on Image Processing}, 
volume={20},
number={8},
pages={2315-2328},
year={2011},
}

@inproceedings{10.1145/2467696.2467741,
author = {Bahrani, Bamdad and Kan, Min-Yen},
title = {Multimodal alignment of scholarly documents and their presentations},
booktitle = ACMMM,
pages = {281–284},
year = {2013}
}

@inproceedings{tp_pres,
  author    = {Tessai Hayama and Hidetsugu Nanba and Susumu Kunifuji},
  title     = {Alignment between a Technical Paper and Presentation Sheets Using a Hidden Markov Model},
  booktitle = {Proceedings of the 9th International Conference on Knowledge-Based Intelligent Information and Engineering Systems (KES 2005)},
  year      = {2005},
  month     = {September},
  pages     = {102--106},
  publisher = {IEEE},
  doi       = {10.1109/AMT.2005.1505278},
  isbn      = {0-7803-9035-0},
  address   = {Melbourne, Australia}
}

@inproceedings{vid-slide-alignment,
  author    = {Xiangyu Wang and Mohan Kankanhalli},
  title     = {Robust Alignment of Presentation Videos with Slides},
  booktitle = {Advances in Multimedia Modeling: 16th International Conference, MMM 2010},
  year      = {2010},
  month     = jan,
  pages     = {311--322},
  publisher = {Springer},
  address   = {Chongqing, China},
  series    = {Lecture Notes in Computer Science},
  volume    = {5916},
  isbn      = {978-3-642-10466-4},
  doi       = {10.1007/978-3-642-10467-1_27}
}

@inproceedings{MaViLS,
  author    = {Katharina Anderer and Andreas Reich and Matthias Wölfel},
  title     = {MaViLS, a Benchmark Dataset for Video-to-Slide Alignment, Assessing Baseline Accuracy with a Multimodal Alignment Algorithm Leveraging Speech, OCR, and Visual Features},
  booktitle = {Proceedings of Interspeech 2024},
  year      = {2024},
  month     = sep,
  pages     = {1375--1379},
  doi       = {10.21437/Interspeech.2024-978},
  publisher = {ISCA},
  address   = {Kos, Greece}
}

@InProceedings{10376585,
  author    = {Lee, Dong Won and Ahuja, Chaitanya and Liang, Paul Pu and Natu, Sanika and Morency, Louis-Philippe},
  title     = {Lecture Presentations Multimodal Dataset: Towards Understanding Multimodality in Educational Videos},
  booktitle = ICCV,
  year      = {2023},
  pages     = {20030--20041},
  doi       = {10.1109/ICCV51070.2023.01838}
}

@InProceedings{10.1145/2647868.2654964,
  author    = {Chen, Huizhong and Cooper, Matthew and Joshi, Dhiraj and Girod, Bernd},
  title     = {Multi-Modal Language Models for Lecture Video Retrieval},
  booktitle = ACMMM,
  year      = {2014},
  pages     = {1081--1084},
  doi       = {10.1145/2647868.2654964}
}

@InProceedings{multicross,
  author    = {Nguyen, Nhu-Van and Coustaty, Micka{\"e}l and Ogier, Jean-Marc},
  title     = {Multi-Modal and Cross-Modal for Lecture Videos Retrieval},
  booktitle = ICPR,
  year      = {2014},
  pages     = {2667--2672},
  doi       = {10.1109/ICPR.2014.461}
}

@InProceedings{doc2ppt,
  author    = {Fu, Tsu-Jui and Wang, William and McDuff, Daniel and Song, Yale},
  title     = {DOC2PPT: Automatic Presentation Slides Generation from Scientific Documents},
  booktitle = AAAI,
  year      = {2022},
  pages     = {634--642},
  doi       = {10.1609/aaai.v36i1.19943}
}

@InProceedings{AutoGenSlide,
  author    = {Cagliero, Luca and La Quatra, Moreno},
  title     = {Automatic Slides Generation in the Absence of Training Data},
  booktitle = {IEEE Annu. Comput. Softw. Appl. Conf. (COMPSAC)},
  year      = {2021},
  pages     = {103--108},
  doi       = {10.1109/COMPSAC51774.2021.00025}
}

@inproceedings{sun2025p2pautomatedpapertopostergeneration,
  author    = {Sun, Tao and Pan, Enhao and Yang, Zhengkai and Sui, Kaixin and Shi, Jiajun and Cheng, Xianfu and Li, Tongliang and Huang, Wenhao and Zhang, Ge and Yang, Jian and Li, Zhoujun},
  title     = {P2P: Automated Paper-to-Poster Generation and Fine-Grained Benchmark},
  booktitle = {Proceedings of the 39th Conference on Neural Information Processing Systems (NeurIPS 2025)},
  year      = {2025},
  url       = {https://arxiv.org/abs/2505.17104}
}

@article{zhu2025paper2videoautomaticvideogeneration,
  author       = {Zeyu Zhu and
                  Kevin Qinghong Lin and
                  Mike Zheng Shou},
  title        = {Paper2Video: Automatic Video Generation from Scientific Papers},
  journal      = {CoRR},
  volume       = {abs/2510.05096},
  year         = {2025},
  url          = {https://doi.org/10.48550/arXiv.2510.05096},
  doi          = {10.48550/ARXIV.2510.05096},
  eprinttype   = {arXiv},
  eprint       = {2510.05096},
  bibsource    = {dblp computer science bibliography, https://dblp.org}
}

@misc{miao2025paper2agentreimaginingresearchpapers,
      title={Paper2Agent: Reimagining Research Papers As Interactive and Reliable AI Agents}, 
      author={Jiacheng Miao and Joe R. Davis and Yaohui Zhang and Jonathan K. Pritchard and James Zou},
      year={2025},
      eprint={2509.06917},
      archivePrefix={arXiv},
      primaryClass={cs.AI},
      url={https://arxiv.org/abs/2509.06917}, 
}

@inproceedings{shi2025presentagentmultimodalagentpresentation,
    title = "{P}resent{A}gent: Multimodal Agent for Presentation Video Generation",
    author = "Shi, Jingwei  and
      Zhang, Zeyu  and
      Wu, Biao  and
      Liang, Yanjie  and
      Fang, Meng  and
      Chen, Ling  and
      Zhao, Yang",
    editor = {Habernal, Ivan  and
      Schulam, Peter  and
      Tiedemann, J{\"o}rg},
    booktitle = "Proceedings of the 2025 Conference on Empirical Methods in Natural Language Processing: System Demonstrations",
    month = nov,
    year = "2025",
    address = "Suzhou, China",
    publisher = "Association for Computational Linguistics",
    url = "https://aclanthology.org/2025.emnlp-demos.58/",
    doi = "10.18653/v1/2025.emnlp-demos.58",
    pages = "760--773",
    ISBN = "979-8-89176-334-0"
}

@inproceedings{frome2013devise,
  title={DeViSE: A Deep Visual-Semantic Embedding Model},
  author = {Frome, Andrea and Corrado, Greg S. and Shlens, Jonathon and Bengio, Samy and Dean, Jeffrey and Ranzato, Marc'Aurelio and Mikolov, Tomas},
  booktitle = {Proceedings of the 27th International Conference on Neural Information Processing Systems - Volume 2},
pages = {2121–2129}, 
numpages = {9},
  year={2013}
}

@inproceedings{radford2021learning,
  title={Learning Transferable Visual Models From Natural Language Supervision},
  author={Radford, Alec and Kim, Jong Wook and Hallacy, Chris and Ramesh, Aditya and Goh, Gabriel and Agarwal, Sandhini and Sastry, Girish and Askell, Amanda and Mishkin, Pamela and Clark, Jack and Krueger, Gretchen and Sutskever, Ilya},
  booktitle={Proceedings of the 38th International Conference on Machine Learning (ICML)},
  year={2021}
}

@inproceedings{alayrac2020self,
  title={Self-Supervised MultiModal Versatile Networks},
  author={Alayrac, Jean-Baptiste and Recasens, Adri{\`a} and Schneider, Rosalia and Arandjelovi{\'c}, Relja and Ramapuram, Jason and De Fauw, Jeffrey and Smaira, Lucas and Dieleman, Sander and Zisserman, Andrew and Carreira, Joao},
  booktitle={Advances in Neural Information Processing Systems},
  year={2020}
}

@inproceedings{miech2019howto100m,
  title={HowTo100M: Learning a Text-Video Embedding by Watching Hundred Million Narrated Video Clips},
  author={Miech, Antoine and Zhukov, Dimitri and Alayrac, Jean-Baptiste and Tapaswi, Makarand and Laptev, Ivan and Sivic, Josef},
  booktitle={Proceedings of the IEEE/CVF International Conference on Computer Vision (ICCV)},
  year={2019}
}

@inproceedings{gabeur2020multi,
  title={Multi-Modal Transformer for Video Retrieval},
  author={Gabeur, Valentin and Sun, Chen and Alahari, Karteek and Schmid, Cordelia},
  booktitle={Proceedings of the European Conference on Computer Vision (ECCV)},
  year={2020}
}

@inproceedings{li2021align,
  title={Align before Fuse: Vision and Language Representation Learning with Momentum Distillation},
  author={Li, Yuan and Lin, Haoxuan and Zhou, Deyao and Zhao, Bin and Guan, Zhiqiang and Wang, Jinqiao and Pu, Shiliang},
  booktitle={Advances in Neural Information Processing Systems},
  year={2021}
}

@inproceedings{kim2021vilt,
  title={ViLT: Vision-and-Language Transformer Without Convolution or Region Supervision},
  author={Kim, Wonjae and Son, Bokyung and Kim, Ildoo},
  booktitle={Proceedings of the 38th International Conference on Machine Learning (ICML)},
  year={2021}
}

@InProceedings{10.1145/2910896.2910904,
  author    = {Clark, Christopher and Divvala, Santosh},
  title     = {PDFFigures 2.0: Mining Figures from Research Papers},
  booktitle = {ACM/IEEE-CS Joint Conf. Digital Libraries (JCDL)},
  year      = {2016},
  pages     = {143--152},
  doi       = {10.1145/2910896.2910904},
  url       = {https://doi.org/10.1145/2910896.2910904}
}

@misc{sun2025ppdoclayoutunifieddocumentlayout,
      title={PP-DocLayout: A Unified Document Layout Detection Model to Accelerate Large-Scale Data Construction}, 
      author={Ting Sun and Cheng Cui and Yuning Du and Yi Liu},
      year={2025},
      eprint={2503.17213},
      archivePrefix={arXiv},
      primaryClass={cs.CV},
      url={https://arxiv.org/abs/2503.17213}, 
}

@inproceedings{mondal-etal-2024-presentations,
  title     = {Presentations by the Humans and For the Humans: Harnessing LLMs for Generating Persona-Aware Slides from Documents},
  author    = {Mondal, Ishani and S, Shwetha and Natarajan, Anandhavelu and Garimella, Aparna and Bandyopadhyay, Sambaran and Boyd-Graber, Jordan},
  booktitle = {Proceedings of the 18th Conference of the European Chapter of the Association for Computational Linguistics (Volume 1: Long Papers)},
  editor    = {Graham, Yvette and Purver, Matthew},
  year      = {2024},
  month     = mar,
  address   = {St. Julian's, Malta},
  publisher = {Association for Computational Linguistics},
  pages     = {2664--2684},
  url       = {https://aclanthology.org/2024.eacl-long.163/},
  doi       = {10.18653/v1/2024.eacl-long.163}
}

@InProceedings{zhang2025gmeimprovinguniversalmultimodal,
    author    = {Zhang, Xin and Zhang, Yanzhao and Xie, Wen and Li, Mingxin and Dai, Ziqi and Long, Dingkun and Xie, Pengjun and Zhang, Meishan and Li, Wenjie and Zhang, Min},
    title     = {Bridging Modalities: Improving Universal Multimodal Retrieval by Multimodal Large Language Models},
    booktitle = {Proceedings of the IEEE/CVF Conference on Computer Vision and Pattern Recognition (CVPR)},
    month     = {June},
    year      = {2025},
    pages     = {9274-9285}
}

@article{DBLP:journals/corr/abs-2407-12580,
  author       = {Ting Jiang and
                  Minghui Song and
                  Zihan Zhang and
                  Haizhen Huang and
                  Weiwei Deng and
                  Feng Sun and
                  Qi Zhang and
                  Deqing Wang and
                  Fuzhen Zhuang},
  title        = {{E5-V:} Universal Embeddings with Multimodal Large Language Models},
  journal      = {CoRR},
  volume       = {abs/2407.12580},
  year         = {2024},
  url          = {https://doi.org/10.48550/arXiv.2407.12580},
  doi          = {10.48550/ARXIV.2407.12580},
  eprinttype   = {arXiv},
  eprint       = {2407.12580},
  bibsource    = {dblp computer science bibliography, https://dblp.org}
}

@inproceedings{attendto2025,
  title={Attend to what I say: Highlighting relevant content on slides},
  author={Megha Mariam K M and C. V. Jawahar},
  booktitle={Proceedings of the International Conference on Document Analysis and Recognition (ICDAR)},
  year={2025}
}

@inproceedings{faysse2024colpali,
  title     = {ColPali: Efficient Document Retrieval with Vision Language Models},
  author    = {Manuel Faysse and Hugues Sibille and Tony Wu and Bilel Omrani and Gautier Viaud and C{\'e}line Hudelot and Pierre Colombo},
  booktitle = {International Conference on Learning Representations (ICLR)},
  year      = {2025}
}

@article{wang2025internvl35advancingopensourcemultimodal,
  author       = {Weiyun Wang and
                  Zhangwei Gao and
                  Lixin Gu and
                  Hengjun Pu and
                  Long Cui and
                  Xingguang Wei and
                  Zhaoyang Liu and
                  Linglin Jing and
                  Shenglong Ye and
                  Jie Shao and
                  Zhaokai Wang and
                  Zhe Chen and
                  Hongjie Zhang and
                  Ganlin Yang and
                  Haomin Wang and
                  Qi Wei and
                  Jinhui Yin and
                  Wenhao Li and
                  Erfei Cui and
                  Guanzhou Chen and
                  Zichen Ding and
                  Changyao Tian and
                  Zhenyu Wu and
                  JingJing Xie and
                  Zehao Li and
                  Bowen Yang and
                  Yuchen Duan and
                  Xuehui Wang and
                  Zhi Hou and
                  Haoran Hao and
                  Tianyi Zhang and
                  Songze Li and
                  Xiangyu Zhao and
                  Haodong Duan and
                  Nianchen Deng and
                  Bin Fu and
                  Yinan He and
                  Yi Wang and
                  Conghui He and
                  Botian Shi and
                  Junjun He and
                  Yingtong Xiong and
                  Han Lv and
                  Lijun Wu and
                  Wenqi Shao and
                  Kaipeng Zhang and
                  Huipeng Deng and
                  Biqing Qi and
                  Jiaye Ge and
                  Qipeng Guo and
                  Wenwei Zhang and
                  Songyang Zhang and
                  Maosong Cao and
                  Junyao Lin and
                  Kexian Tang and
                  Jianfei Gao and
                  Haian Huang and
                  Yuzhe Gu and
                  Chengqi Lyu and
                  Huanze Tang and
                  Rui Wang and
                  Haijun Lv and
                  Wanli Ouyang and
                  Limin Wang and
                  Min Dou and
                  Xizhou Zhu and
                  Tong Lu and
                  Dahua Lin and
                  Jifeng Dai and
                  Weijie Su and
                  Bowen Zhou and
                  Kai Chen and
                  Yu Qiao and
                  Wenhai Wang and
                  Gen Luo},
  title        = {InternVL3.5: Advancing Open-Source Multimodal Models in Versatility,
                  Reasoning, and Efficiency},
  journal      = {CoRR},
  volume       = {abs/2508.18265},
  year         = {2025},
  url          = {https://doi.org/10.48550/arXiv.2508.18265},
  doi          = {10.48550/ARXIV.2508.18265},
  eprinttype   = {arXiv},
  eprint       = {2508.18265},
  bibsource    = {dblp computer science bibliography, https://dblp.org}
}

@article{Anmarkrud03042019,
author = {Øistein Anmarkrud, Anette Andresen and Ivar Bråten},
title = {Cognitive Load and Working Memory in Multimedia Learning: Conceptual and Measurement Issues},
journal = {Educational Psychologist},
volume = {54},
number = {2},
pages = {61--83},
year = {2019},
publisher = {Routledge},
doi = {10.1080/00461520.2018.1554484},
URL = {    
        https://doi.org/10.1080/00461520.2018.1554484
},
eprint = { 
        https://doi.org/10.1080/00461520.2018.1554484
}
}

@inproceedings{Cog2, 
author = {Plass, Jan L. and Homer, Bruce D.},
title = {Cognitive Load in Multimedia Learning: The Role of Learner Preferences and Abilities},
year = {2002},
isbn = {0769515096},
publisher = {IEEE Computer Society},
address = {USA},
booktitle = {Proceedings of the International Conference on Computers in Education},
pages = {564},
series = {ICCE '02}
}

@article{qwen2025qwen25technicalreport,
  author       = {An Yang and
                  Baosong Yang and
                  Beichen Zhang and
                  Binyuan Hui and
                  Bo Zheng and
                  Bowen Yu and
                  Chengyuan Li and
                  Dayiheng Liu and
                  Fei Huang and
                  Haoran Wei and
                  Huan Lin and
                  Jian Yang and
                  Jianhong Tu and
                  Jianwei Zhang and
                  Jianxin Yang and
                  Jiaxi Yang and
                  Jingren Zhou and
                  Junyang Lin and
                  Kai Dang and
                  Keming Lu and
                  Keqin Bao and
                  Kexin Yang and
                  Le Yu and
                  Mei Li and
                  Mingfeng Xue and
                  Pei Zhang and
                  Qin Zhu and
                  Rui Men and
                  Runji Lin and
                  Tianhao Li and
                  Tingyu Xia and
                  Xingzhang Ren and
                  Xuancheng Ren and
                  Yang Fan and
                  Yang Su and
                  Yichang Zhang and
                  Yu Wan and
                  Yuqiong Liu and
                  Zeyu Cui and
                  Zhenru Zhang and
                  Zihan Qiu},
  title        = {Qwen2.5 Technical Report},
  journal      = {CoRR},
  volume       = {abs/2412.15115},
  year         = {2024},
  url          = {https://doi.org/10.48550/arXiv.2412.15115},
  doi          = {10.48550/ARXIV.2412.15115},
  eprinttype   = {arXiv},
  eprint       = {2412.15115},
  bibsource    = {dblp computer science bibliography, https://dblp.org}
}

@inproceedings{wang2024slideavsrdatasetpaperexplanation,
    title = "{S}lide{AVSR}: A Dataset of Paper Explanation Videos for Audio-Visual Speech Recognition",
    author = "Wang, Hao  and
      Kurita, Shuhei  and
      Shimizu, Shuichiro  and
      Kawahara, Daisuke",
    editor = "Gu, Jing  and
      Fu, Tsu-Jui (Ray)  and
      Hudson, Drew  and
      Celikyilmaz, Asli  and
      Wang, William",
    booktitle = "Proceedings of the 3rd Workshop on Advances in Language and Vision Research (ALVR)",
    month = aug,
    year = "2024",
    address = "Bangkok, Thailand",
    publisher = "Association for Computational Linguistics",
    url = "https://aclanthology.org/2024.alvr-1.11/",
    doi = "10.18653/v1/2024.alvr-1.11",
    pages = "129--137"
}

@inproceedings{WangHG25-2,
  title = {DocVideoQA: Towards Comprehensive Understanding of Document-Centric Videos through Question Answering},
  author = {Haochen Wang and Kai Hu and Liangcai Gao},
  year = {2025},
  doi = {10.1109/ICASSP49660.2025.10887668},
  url = {https://doi.org/10.1109/ICASSP49660.2025.10887668},
  researchr = {https://researchr.org/publication/WangHG25-2},
  cites = {0},
  citedby = {0},
  pages = {1-5},
  booktitle = {2025 IEEE International Conference on Acoustics, Speech and Signal Processing, ICASSP 2025, Hyderabad, India, April 6-11, 2025},
  publisher = {IEEE},
  isbn = {979-8-3503-6874-1},
}

@article{DBLP:journals/corr/abs-2502-20582,
  author       = {Javin Liu and
                  Aryan Vats and
                  Zihao He},
  title        = {CS-PaperSum: {A} Large-Scale Dataset of AI-Generated Summaries for
                  Scientific Papers},
  journal      = {CoRR},
  volume       = {abs/2502.20582},
  year         = {2025},
  url          = {https://doi.org/10.48550/arXiv.2502.20582},
  doi          = {10.48550/ARXIV.2502.20582},
  eprinttype   = {arXiv},
  eprint       = {2502.20582},
  bibsource    = {dblp computer science bibliography, https://dblp.org}
}

@article{DBLP:journals/corr/abs-2009-09941,
  author       = {Yuning Du and
                  Chenxia Li and
                  Ruoyu Guo and
                  Xiaoting Yin and
                  Weiwei Liu and
                  Jun Zhou and
                  Yifan Bai and
                  Zilin Yu and
                  Yehua Yang and
                  Qingqing Dang and
                  Haoshuang Wang},
  title        = {{PP-OCR:} {A} Practical Ultra Lightweight {OCR} System},
  journal      = {CoRR},
  volume       = {abs/2009.09941},
  year         = {2020},
  url          = {https://arxiv.org/abs/2009.09941},
  eprinttype   = {arXiv},
  eprint       = {2009.09941},
  bibsource    = {dblp computer science bibliography, https://dblp.org}
}

@inproceedings{sun-etal-2021-d2s,
    title = "{D}2{S}: Document-to-Slide Generation Via Query-Based Text Summarization",
    author = "Sun, Edward  and
      Hou, Yufang  and
      Wang, Dakuo  and
      Zhang, Yunfeng  and
      Wang, Nancy X. R.",
    editor = "Toutanova, Kristina  and
      Rumshisky, Anna  and
      Zettlemoyer, Luke  and
      Hakkani-Tur, Dilek  and
      Beltagy, Iz  and
      Bethard, Steven  and
      Cotterell, Ryan  and
      Chakraborty, Tanmoy  and
      Zhou, Yichao",
    booktitle = "Proceedings of the 2021 Conference of the North American Chapter of the Association for Computational Linguistics: Human Language Technologies",
    month = jun,
    year = "2021",
    address = "Online",
    publisher = "Association for Computational Linguistics",
    url = "https://aclanthology.org/2021.naacl-main.111/",
    doi = "10.18653/v1/2021.naacl-main.111",
    pages = "1405--1418"
}

@inproceedings{SlidesGen,
  author    = {Sravanthi, M. and Chowdary, C. R. and Kumar, P.},
  title     = {SlidesGen: Automatic Generation of Presentation Slides for a Technical Paper Using Summarization},
  booktitle = {Proceedings of the International Conference on Intelligent Agent \& Multi-Agent Systems (IAMA)},
  year      = {2009},
  publisher = {IEEE},
  doi       = {10.1109/IAMA.2009.5228050},
  url       = {https://doi.org/10.1109/IAMA.2009.5228050}
}
